# Probabilistic Neural Network Training for Semi-Supervised Classifiers

*Hamidreza Farhidzadeh

*Department of Mathematicss and Computer Science, Amirkabir University of Technology, Tehran, Iran

**Abstract**: In this paper, we propose another version of help-training approach by employing a Probabilistic Neural Network (PNN) that improves the performance of the main discriminative classifier in the semi-supervised strategy. We introduce the PNN-training algorithm and use it for training the support vector machine (SVM) with a few numbers of labeled data and a large number of unlabeled data. We try to find the best labels for unlabeled data and then use SVM to enhance the classification rate. We test our method on two famous benchmarks and show the efficiency of our method in comparison with pervious methods.

## 1- Introduction

Traditionally, there were two main strategies to learn classifiers: supervised and unsupervised learning. In the supervised learning, classifiers use only labeled data to train. Obtaining the labeled data is difficult, expensive, and time consuming because it needs human knowledge and ability. One the other hand, in the unsupervised learning, there is no supervisor to learn classifiers to label the data. In this strategy, classifier work with only unlabeled data and classify them according to some similar features they have. In the semi-supervised learning method, which had been introduced by M. Seeger [11], has been found that unlabeled data, when used in conjunction with a small amount of labeled data, can produce considerable improvement in learning accuracy. Because semi-supervised learning requires less human effort and gives higher accuracy, it is of great interest both in theory and in practice.

There are different methods in supervised learning such as Expectation-Maximization (EM) algorithm [1], Co-Training [4], Transductive Support Vector Machine (TSVM) [8], Self-Training [14] and Help-Training [2].

In Self-Training, classification process is done in two stages. First, the classifier is trained by known data and then it uses to predict the labels of unlabeled data. The unlabeled data with high confident scores adds to the training set with estimated labels, and then this process repeats until the convergence. This semi-supervised strategy has a problem with discriminative classifiers and the samples will be classified in the wrong classes with most confident scores.

The Help-Training method is an improved version of the Self-Training classical technique. In this method, a generative model, *G,* helps the discriminative classifier, *C*, to decide about samples that can be labeled and conjunct to the training set. In each level, *G* finds data that obtain a high probability in order to belong to each class, and classifier *C* classifies these data which are selected and added by a G training set. This process accomplishes until there is no unlabeled data.

In this paper, we present another version of the Help-Training strategy by using different method. In [2] Parzen window estimator is used as *G*, but we used Parzen Probabilistic Neural Network (PPNN) or in short term Probabilistic Neural Network. The rest of this article is organized as follows. Section 2 presents the PNN. Section 3 reviews the SVM classifier. Section 4 provides the PNN-Training algorithm and gives an application of this algorithm to SVM. Finally, we present our experimental results and conclusion in Section 5 and 6, respectively.

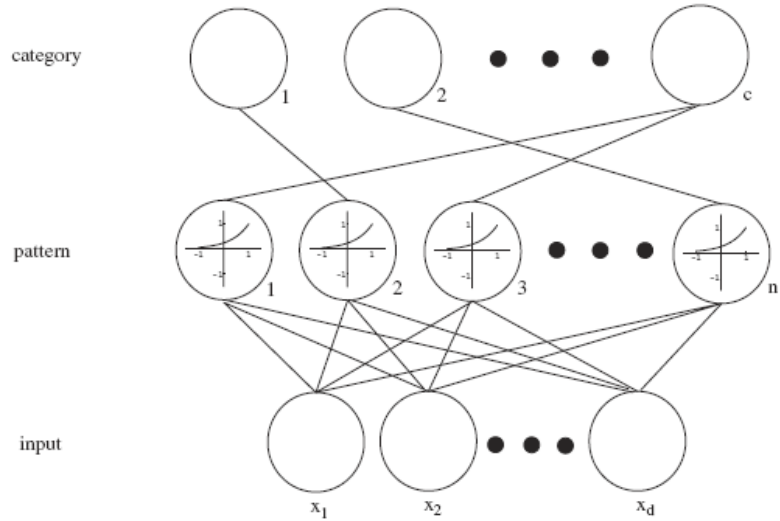

**Figure 1. PNN structure**

### 2- PNN

A PNN is a feed forward neural network, which was derived from Bayesian network and a statistical algorithm called Kernel Fisher discriminant analysis [10]. It was introduced by D.F. Specht [13].

PNN made three assumptions: if each classification probability density function has the same type, then it is the Gaussian distribution and it is also the normal distribution. Each classification which Gaussian distribution probability density function of the covariance matrix is diagonal matrix and each values is the same [12].

A PNN consists of $d$ input units, $n$ pattern units and $c$ category units. Each pattern unit forms the inner product of its weight vector and the normalized pattern vector $x$ to form $z = w^t x$, and then emits $exp[(z-1)/\sigma^2]$, where $\sigma$ is a parameter set by the user and is equal to $\sqrt{2}$ times the width of the effective Gaussian window . Each category unit sums such contributions from the pattern unit connected to it. This insures that the activity in each of the category units represents the Parzen window density estimate using a circularly symmetric Gaussian window of covariance $\sigma^2 I$, where $I$ is the $d \times d$ identity matrix. The structure of PNN is shown in Fig.1. If we denote the components of the $j$-th pattern as $x_{jk}$ and the weights to the $j$-th pattern unit $w_{jk}$, for $j = 1, 2,..., n$ and $k = 1, 2, ..., d$, then PNN training algorithm is:

**PNN Training Algorithm**
    1) Begin initialize $j = 0$, $n =$ number of patterns
    2)         Do $j = j + 1$

3)                        Normalize: $x_{jk} \leftarrow x_{jk} / (\sum_{i}^{d} x_{ji}^2)^{1/2}$

4)                        Train: $w_{jk} \leftarrow x_{jk}$

5)                        If $x \in w_i$ then $a_{ic} \leftarrow 1$

6)                        Until $j = n$

7) End

For the classification of pattern *x* is as follows: Each pattern is placed at the input units. Each pattern unit computes the inner product:

$$z_k = w_k^t x,$$

and emits a nonlinear function of $z_k$. Each pattern unit contributes to its associated category unit a signal equal to the probability the test point was generated by a Gaussian centered on the associated training point. The sum of these local estimates (computed at the corresponding category unit) gives the discriminant function $g_i(x)$ — the Parzen window estimate of the underlying distribution. The *max $g_i(\mathbf{x})$* operation gives the desired category for the test point.

**PNN Classification Algorithm**

1) Begin initialize $k = 0$, x = test pattern
2)                 Do $k \leftarrow k + 1$
3)                 $z_k = w_k^t x$
4)                 If $a_{kc} = 1$ Then $g_c \leftarrow g_c + \exp[(z_k - 1)/\sigma^2]$
5)                 Until $k = n$
6)                 Return $class \leftarrow \arg\max_i g_i(x)$
7) End

There are two significant features of PNN; since the learning rule is simple and requires only a single pass through the training data, the speed of learning is high. Another advantage is that new training patterns can be incorporated quite easily into a previously trained classifier.

    **3- SVM**

Support Vector Machines (SVMs) are a class of supervised learning algorithms introduced by Vapnik[7]. Given a set of labeled training vectors, SVMs learn a linear decision boundary to discriminate between the two classes. Let consider a binary classification problem and a dataset $\{(x_1, y_1), ..., (x_k, y_k)\}$ with $x_i \in R^d$ and $y_i \in \{-1, 1\}$. In the feature space, the decision function is:

$$f(x) = sign(<w, x> + b)$$

where *w* is a normal vector to the hyper plane and *b* is the bias. The *w* and *b* are found by solving this optimization problem which expresses the maximization of the margin $1/\|w\|$ and the minimization of the training error:

$$\min_{w,\xi} \frac{1}{2} w'w + C \sum_{i=1}^{l} \xi_i$$

$$s.t.: y_i[<w,x>+b] \geq 1-\xi_i \quad \forall i=1,...,l$$

$$\xi_i \geq 0 \quad \forall i=1,...,l$$

A key feature of any SVM optimization problem is that it is equivalent to solving a dual quadratic programming problem. In the linearly separable case, the maximal margin classifier is found by solving for the optimal "weights", $\alpha_i$, $i= 1,..., n$ in the dual problem:

$$\underset{\alpha,b}{Max} \sum_{i=1}^{l} \alpha_i - \frac{1}{2} \sum_{i,j=1}^{l} \alpha_i \alpha_j y_i y_j <x_i, x_j>$$

$$s.t.: \sum_{i=1}^{l} \alpha_i y_i = 0 \quad and \quad 0 \leq \alpha_i \leq C, i=1,...,l$$

For nonlinear problems, SVMs use the kernel trick to produce nonlinear boundaries. The idea behind kernels is to map training data nonlinearly into a higher-dimensional feature space via a mapping function $\Phi$ and to construct a separating hyperplane that maximizes the margin. The construction of the linear decision surface in this feature space requires only the evaluation of dot products $\Phi(x_i).\Phi(x_j) = k(x_i, x_j)$, where $k()$ is called the kernel function [5].

## 4- PNN-Training Algorithm

This algorithm is another version of the Help-Training algorithm. In this algorithm, a generative classifier, PNN, models the whole data space and then discriminative classifier, for example SVM, classifies the data.

In data space, we have a few numbers of labeled data. First, we train the PNN with this group of data and then predict the labels of unlabeled data. In this level, the PNN assigns the unlabeled data to one class based on their probability to belong to each class. Then this group of data will be added to the training set. The discriminative classifier classifies the labeled and new labeled data which the PNN estimated belonging of each sample to each class. The PNN-Training for semi-supervised SVM is described as follows:

**PNN-Training for semi-supervised SVM**

1) Input: Set L = Labeled data, U = Unlabeled data
2) Train PNN by L (PNN Training Algorithm)
3) Classify U by trained PNN (PNN Classification Algorithm)
4) Add new labeled data to training set
5) Make a model of training set by SVM
6) Calculate SVM test error by testing set
7) Output: Error classification rate

## 5- Experimental Result

In this section, we test our algorithm on two specific data sets that are common in semi-supervised testing. In this section, we use LIBSVM library [6].

### 5-1 **Two moon**

One of the famous benchmarks in testing of semi-supervised learning algorithm is the Two Moons data set [3, 7].
We produce 50 samples of each class and select randomly 10 samples of each class as a labeled data. After training of PPN by 20% of whole data and estimating of other samples, the training set is ready to be trained and test by SVM. We use RBF kernel with default values of its terms in LIBSVM; gamma = (# feature)$^{-1}$.

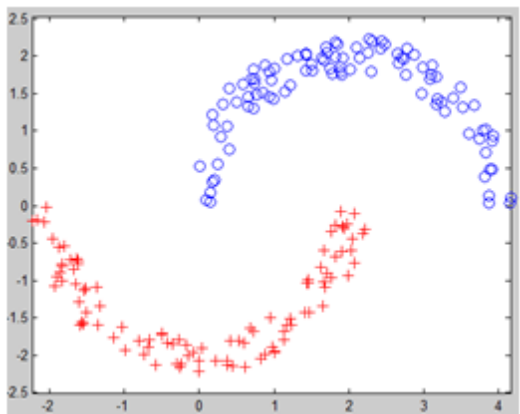

Figure 2.Two moons dataset

The results are shown in Table I. The other test errors are from [2]:

| Method | Test error (%) |
|---|---|
| Branch and Bound | 0 |
| Self-Training | 33.68 |
| Help-Training | 15.07 |
| PNN-Training | 10.23 |

The PNN-Training result shows that it can have better performance that it's previous version, Help-Training. Branch and Bound Algorithm has best efficiency, however, it loses its performance on large datasets.

5-2 **USPS Dataset**
The US Postal Service handwritten digits recognition corpus (USPS) is another famous benchmark to test semi-supervised algorithm. It contains normalized grey scale images of size 16×16 digits, divided into a training set of 7291 images and a test set of 2007 images. 90% of samples are unlabeled. Again we apply a default value of RBF kernel in LIBSVM for supervised SVM. . The results are as following in Table II. The other test errors are from [2]:

| Method | Test error (%) |
|---|---|
| TSVM/SVMlight | 14.70 |
| Self-Training | 8.22 |
| Help-Training | 7.77 |
| PNN-Training | 7.42 |

As we have 10 classes (10 digits) we should use one-against-all strategy instead of pairwise strategy, because we know just 10% of data labels. LIBSVM is a powerful tool to implement this strategy. We compare our result with TSVM which is implemented in SVMlight.[9]. The PNN-Training method can obtain a slightly better classification rate than its previous version, Help-Training method.

### 6- Conclusion

In this paper, we introduced a new type of Help-Training algorithm and name it PNN-Training. We employed this method to one discriminative classifier, SVM. The algorithm estimates the label of large numbers of unlabeled samples by using a few labeled numbers of them. Whole labeled and new estimated labeled samples contribute the main training set. Finally, the RBF-SVM classifier classifies them. We tested our method on two data sets and showed it had better performance in one data set than Help-Training and other semi-supervised learning methods such as TSVM.